\documentclass[runningheads,a4paper]{llncs}

\usepackage{amssymb}
\usepackage{graphicx}
\usepackage[hidelinks]{hyperref}
\usepackage{mathtools}
\usepackage{subcaption}

\usepackage{url}
\urldef{\mailsa}\path|j_borges15@rhrk.uni-kl.de|

\newcommand{\keywords}[1]{\par\addvspace\baselineskip
\noindent\keywordname\enspace\ignorespaces#1}

\begin{document}

\mainmatter  

\title{Deep Learning for Time-Series Analysis}


%
%
\author{John Gamboa%
}

\authorrunning{Deep Learning for Time-Series Analysis}


\institute{University of Kaiserslautern\\
Kaiserslautern, Germany}

%
%

\toctitle{Lecture Notes in Computer Science}
\tocauthor{Authors' Instructions}
\maketitle

\begin{abstract}
In many real-world application, e.g., speech recognition or sleep stage
classification, data are captured over the course of time, constituting a
Time-Series. Time-Series often contain temporal
dependencies that cause two otherwise identical points of time to belong to
different classes or predict different behavior. This characteristic generally
increases the difficulty of analysing them. Existing techniques often
depended on hand-crafted features that were expensive to create and required expert
knowledge of the field. With the advent of Deep Learning new models of
unsupervised learning of features for Time-series analysis and forecast have
been developed. Such new developments are the topic of this paper: a review of
the main Deep Learning techniques is presented, and some applications on
Time-Series analysis are summaried. The results make it clear that Deep Learning
has a lot to contribute to the field.

\end{abstract}

\section{Introduction} \label{secIntroduction}

Artificial Neural Networks (ANN), since their origin in 1943
\cite{mcculloch1943logical},
have been used to solve a large range of problems as diverse as
robotic processing \cite{king1989neural}, object recognition
\cite{szegedy2013deep}, speech and handwriting recognition
\cite{graves2012supervised}, and even real time sign-language translation
\cite{akmeliawati2007real}. Despite the intuition that deeper architectures
would yield better results than the then more commonly used shallow ones,
empirical tests with deep networks had found similar or even worse results when
compared to networks with only one or two layers
\cite{tesauro1992practical} (for more details, see \cite{bengio2007greedy}).
Additionally, training was found to be difficult and often inefficient
\cite{bengio2007greedy}. L{\"a}ngkvist \cite{langkvist2014modeling} argues that
this scenario started to change with the proposal of
{\em greedy layer-wise unsupervised learning} \cite{hinton2006fast}, which
allowed for the fast learning of Deep Belief Networks, while also solving the
vanishing gradients problem \cite{bengio1994learning}. Latest deep architectures
use several modules that are trained separately and stacked together so that the
output of the first one is the input of the next one.

From stock market prices to the spread of an epidemic, and from the recording
of an audio signal to sleep monitoring, it is common for real world data to be
registered taking into account some notion of time. When collected together,
the measurements compose what is known as a {\em Time-Series}. For different
fields, suitable applications vary depending on the nature and purpose of the
data: while doctors can be interested in searching for anomalies in the sleep
patterns of a patient, economists may be more interested in forecasting the next
prices some stocks of interest will assume.
These kinds of problems are addressed in the literature by a range of different
approches (for a recent review of the main techniques applied to perform tasks
such as Classification, Segmentation, Anomaly Detection and Prediction, see
\cite{esling2012time}).

This paper reviews some of the recently presented approaches to performing tasks
related to Time-Series using Deep Learning architectures. It is important,
therefore, to have a formal definition of Time-Series.
Malhotra et al. \cite{malhotra23long} defined Time-Series as a vector 
$X = \{\textbf{x}^{(1)}, \textbf{x}^{(2)}, \dots, \textbf{x}^{(n)}\}$, where
each element $\textbf{x}^{(t)} \in R^m$ pertaining to $X$ is an array of $m$
values such that $\{ x_1^{(t)}, x_2^{(t)}, \dots, x_m^{(t)} \}$. Each one of
the $m$ values correspond to the input variables measured in the time-series.

The rest of this paper is structured as follows:
Section \ref{secNeuralNetworks}
introduces basic types of Neural Network (NN) modules that are often used to build
deep neural structures.
Section \ref{secRelatedWork} describes how the present
paper relates to other works in the literature. Sections
\ref{secDLModeling}, \ref{secDLClassification} and \ref{secDLAnomalyDetect}
describe some approaches using Deep Learning to perform Modeling, Classification
and Anomaly Detection in Time-Series data, respectively. Finally, Section
\ref{secConclusion} concludes the paper.





\subsection{Artificial Neural Network} \label{secNeuralNetworks}

This section explains the basic concepts related to ANN. The types of networks
described here are by no means the only kinds of ANN architectures found in the
literature. The reader is referred to \cite{rojas2013neural} for a thorough
description of architectural alternatives such as Restricted Boltzmann Machines
(RBM), Hopfield Networks and Auto-Encoders, as well as for a detailed
explanation of the Backpropagation algorithm. Additionally, we refer the reader
to  \cite{graves2012supervised} for applications of RNN as well as more details
on the implementation of a LSTM, and to \cite{stutz2014understanding} for
details on CNN.

An ANN is basically a network of computing units linked by
directed connections. Each computing unit performs some calculation and outputs
a value that is then spread through all its outgoing connections as input into
other units. Connections normally have weights that correspond to how strong two
units are linked. Typically, the computation performed by a unit is separated
into two stages: the {\em aggregation} and the {\em activation} functions.
Applying the aggregation function commonly corresponds to calculating the
sum of the inputs received by the unit through all its incoming connections.
The resulting value is then fed into the activation function. It commonly
varies in different network architectures,
although popular choices are the logistic sigmoid
($\sigma(x) = \frac{1}{1 + e^{-x}}$) and the hyperbolic tangent
($tanh(x) = \frac{2}{1 + e^{-2x}} -1$) functions. Recently, rectified linear
units employing a ramp function
($R(x) = \mathrm{max}(0, x)$) have become increasingly popular.

\begin{figure}
\centering
\begin{subfigure}{.33\textwidth}
  \centering
  \includegraphics[width=1.1\linewidth]{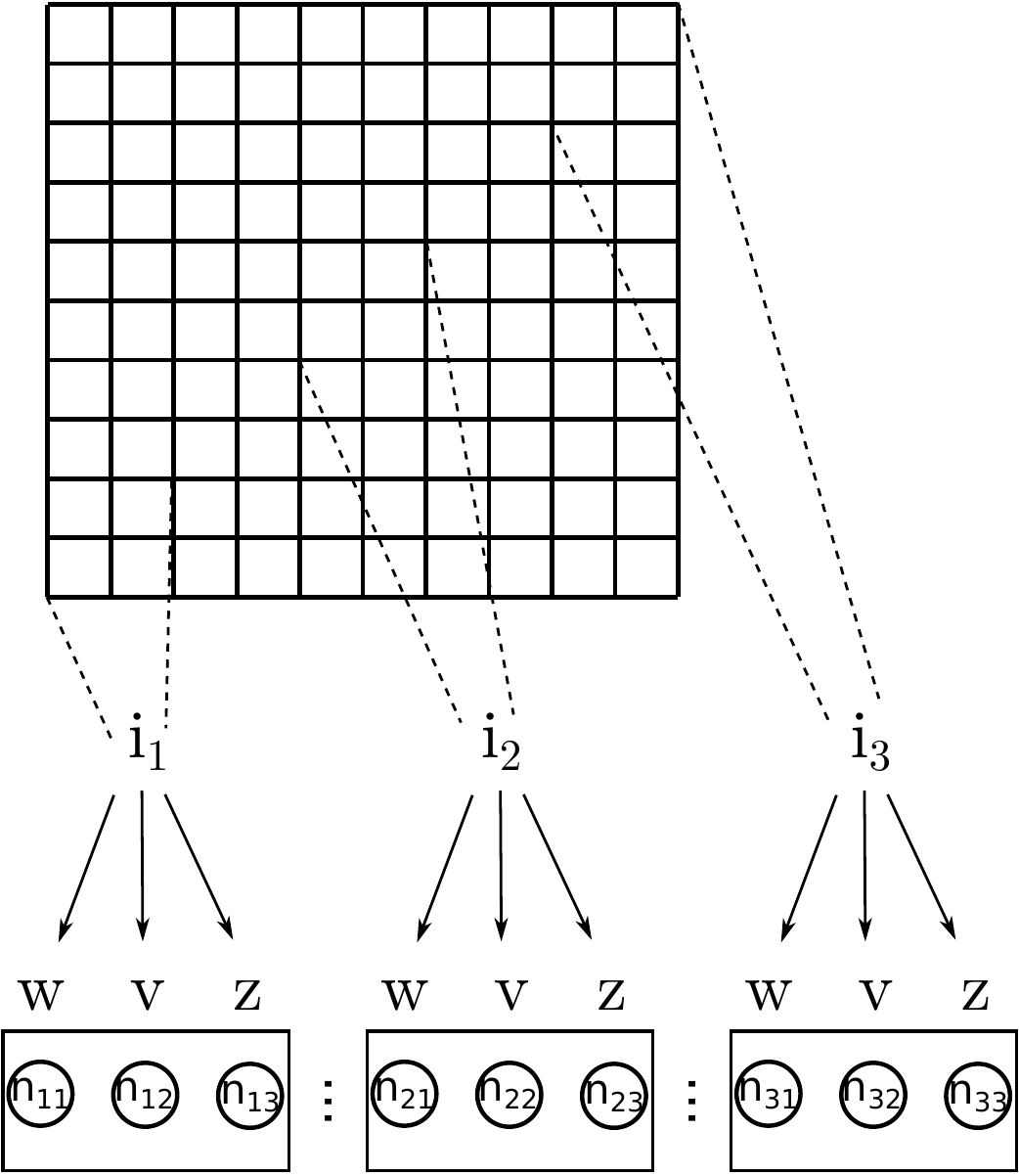}
  \caption{}
  \label{figCNN}
\end{subfigure}
\begin{subfigure}{.33\textwidth}
  \centering
  \includegraphics[width=0.7\linewidth]{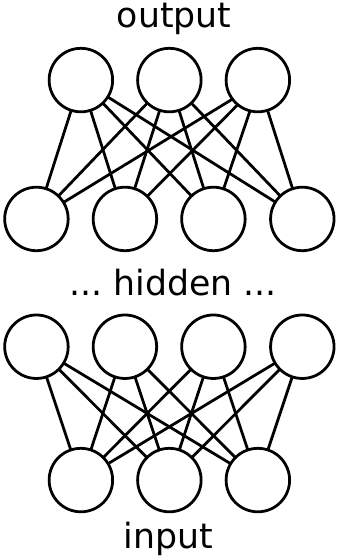}
  \caption{}
  \label{figANN}
\end{subfigure}%
\begin{subfigure}{.33\textwidth}
  \centering
  \includegraphics[width=1\linewidth]{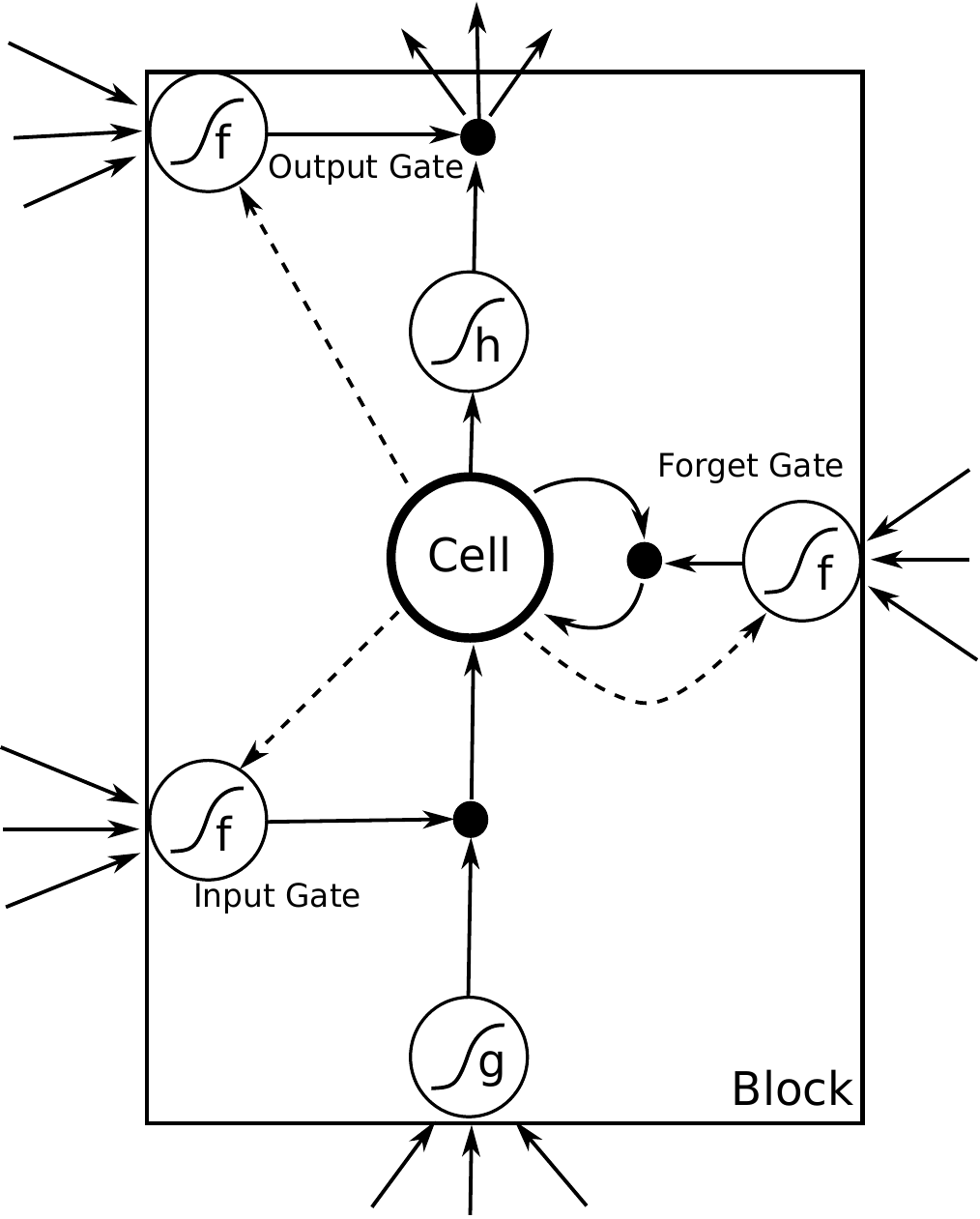}
  \caption{}
  \label{figLSTM}
\end{subfigure}
\caption{(a) The {\em convolutional layer} of a CNN with three groups
(also called ``filters''). Each
group performs a $2 \times 2$ convolution in the image: each neuron in the group
is connected to a different region of the image but shares the same weights,
producing a new image.
In the example, the weight vector $w$ is shared by all neurons $n_{j1}$, the
vector $v$ is shared by all neurons $n_{j2}$, and $z$ is shared by all neurons
$n_{j3}$. If pooling is applied, it is applied to each one of the three newly
produced images;
(b) An Artificial Neural Network with one input layer composed of three
neurons, two hidden layers composed, each one, of four neurons, and one output
layer composed of three neurons. Each node of a layer is connected to all nodes
of the next layer;
(c) A LSTM block (adapted from \protect\cite{graves2012supervised}).}
\label{figExamples}
\end{figure}

The input of the network is given in a set of input computing units which
compose an {\em input layer}. Conversely, the output of the network are
the values output by the units composing the {\em output layer}. All other units
are called {\em hidden} and are often also organized in layers (see Figure
\ref{figANN} for an example network).

\begin{figure}
\centering
\begin{subfigure}{.33\textwidth}
  \centering
  \includegraphics[width=0.7\linewidth]{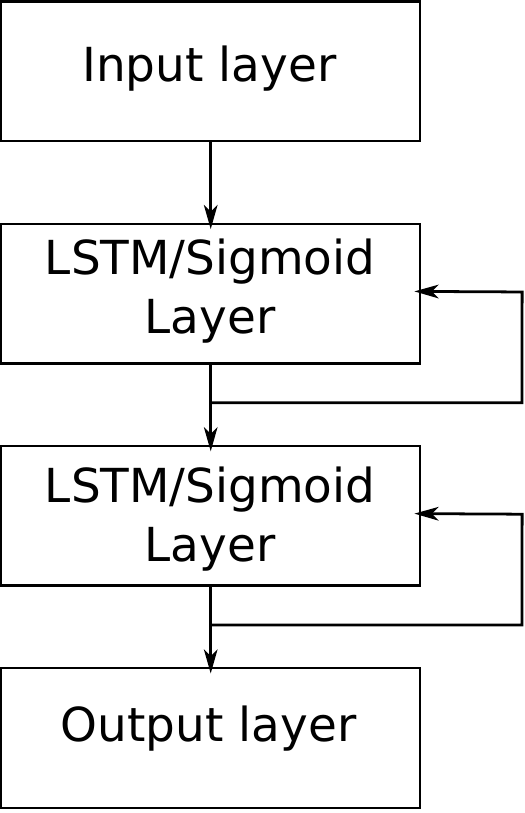}
  \caption{}
  \label{figLSTMAD}
\end{subfigure}
\begin{subfigure}{.66\textwidth}
  \centering
  \includegraphics[width=1\linewidth]{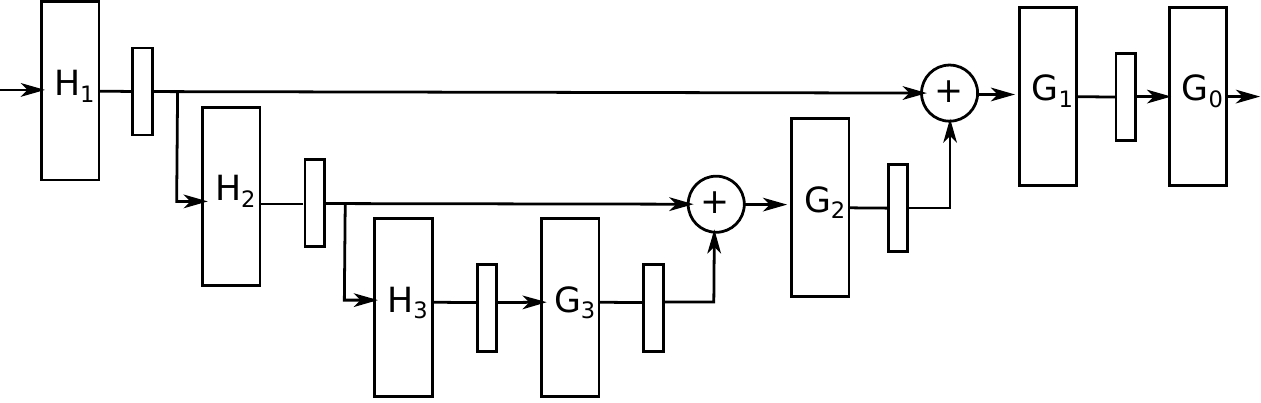}
  \caption{}
  \label{figUFCNN}
\end{subfigure}%
\caption{(a) The proposed ``Stacked Architecture'' for performing Anomaly
Detection (adapted from \protect\cite{malhotra23long}); (b) The architecture of a UFCNN
(adapted from \protect\cite{mittelman2015time}).}
\label{figArchitectures}
\end{figure}

The focus of learning algorithms is frequently on deciding what weights would
cause the network to output, given some input, the expected values.
A popular learning algorithm is the Backpropagation algorithm
\cite{rumelhart1988learning}, whereby the gradient of an error function is
calculated and the weights are iteratively set so as to minimize the error.

\subsubsection{Convolutional Neural Network (CNN)} \label{secCNN}

A network that is too big and with layers that are fully connected can become
infeasible to train. Also trained with the Backpropagation algorithm, CNNs
\cite{lecun1998gradient} are common for image processing tasks and reduce the
number of parameters to be learned by limiting the number of connections of the
neurons in the hidden layer to only some of the input neurons (i.e., a local
area of the input image). A hidden layer (in this case, also called a {\em
convolutional layer} -- see Figure \ref{figCNN}) is composed by several groups
of neurons. The weights of all neurons in a group are shared. Each group is
generally composed by as many neurons as needed to cover the entire image. This
way, it is as if each group of neurons in the hidden layer calculated a
convolution of the image with their weights, resulting in a ``processed'' version
of the image. We call this convolution a {\em feature}.

Commonly, {\em pooling} is applied to the resulting filtered images. The
tecnique allows for achieving some translation invariance of the learned
features. The groups (containing a newly processed version of the input
image) are divided in chunks (e.g., $2 \times 2$) and their maximum value is
taken. This results in yet another version of the input image, now smaller than
the original size (in the example, $1/4$ of the size of the group).

These steps can be repeatedly applied as many times as desired: a new
convolutional layer can be applied on the pooled layer, followed by another
pooling layer, and so forth. Finally, when the layers become small enough, it is
common to have fully connected layers before the output layer.

\paragraph{Tiled Convolutional Neural Network}

The usage of shared weights in a group allow for the translation invariance of
the learned features. However, ``it prevents the pooling units from capturing
more complex invariances, such as scale and rotation invariance''
\cite{ngiam2010tiled}. To solve this problem, Tiled CNNs allow for a group to
be divided into subgroups called {\em tiles}, each of which can have separate
weights. A parameter {\em k} defines how many tiles each group has: neurons that
are exactly {\em k} steps away from each other share the same weights.

\paragraph{Fully Convolutional Networks (FCN)}

While the pooling operation performed by CNNs makes sense for object recognition
tasks, because it has the advantage of achieving some robustness to small shifts
of the learned features, it is not suited for tasks like Semantic Segmentation,
where the goal is to segment the pixels of the image according to the objects
that they refer to. FCNs \cite{long2014fully}
allow for the input and output layers to have the same dimensions by introducing
``a decoder stage that is consisted of upsampling, convolution, and rectified
linear units layers, to the CNN architecture'' \cite{mittelman2015time}.

\subsubsection{Recurrent Neural Network (RNN)}

When the network has loops, it is called a RNN. It
is possible to adapt the Backpropagation algorithm to train a recurrent network,
by ``unfolding'' the network through time and constraining some of the connections
to always hold the same weights \cite{rumelhart1988learning}.


\paragraph{Long Short-Term Memory (LSTM)}

One problem that arises from the unfolding of an RNN is that the gradient of
some of the weights starts to become too small or too large if the network is
unfolded for too many time steps. This is called the {\em vanishing gradients}
problem \cite{bengio1994learning}. A type of network architecture that solves
this problem is the LSTM \cite{hochreiter1997long}. In
a typical implementation, the hidden layer is replaced by a complex block
(see Figure \ref{figLSTM}) of
computing units composed by gates that trap the error in the block, forming a
so-called ``error carrousel''.

\section{Literature Review} \label{secRelatedWork}

Independently of Deep Learning, analysis of Time-Series data have been a popular
subject of interest in other fields such as Economics, Engineering and Medicine.
Traditional techniques on manipulating such data can be found in
\cite{hamilton1994time}, and the application of traditional ANN techniques on
this kind of data is described in \cite{azoff1994neural}.

Most work using ANN to manipulate Time-Series data focuses on modeling and
forecasting. As an early attempt on using ANN for such tasks,
\cite{chakraborty1992forecasting} modelled flour prices over the range
of 8 years. Still in the 90's, \cite{kaastra1996designing} delineated eight
steps on ``designing a neural network forecast model using economic time series
data''. More recent approaches include usage of Elman RNNs to predict
chaotic Time-Series \cite{chandra2012cooperative}, employing ANN ensemble methods
for forecasting Internet traffic \cite{cortez2012multi}, using simple Multilayer
Perceptrons for modeling the amount of littering in the North Sea
\cite{schulz2014artificial}, and implementing in FPGA a prediction algorithm
using Echo State Networks for ``exploiting the inherent parallelism of these
systems'' \cite{alomar2015fpga}.

Hybrid approaches to Time-Series analysis utilizing ANN are not uncommon.
\cite{khashei2011novel} presents a model for Time-Series forecasting using ANN
and ARIMA models, and \cite{faruk2010hybrid} applies the same kinds of models
to water quality time series prediction. In still other examples of the same
ideas, \cite{kardakos2013application} compares the performance of ARIMA models
and ANNs to make short-term predictions on photovoltaic power generators, while
\cite{lin2011forecasting} compares both models with the performance of
Multivariate Adaptive Regression Splines. \cite{egrioglu2013fuzzy} performs
Time-Series forecasting by using a hybrid fuzzy model: while the Fuzzy C-means
method is utilized for fuzzification, ANN are employed for defuzzification.
Finally, \cite{hu2013hybrid} forecasts the speed of the wind using a hybrid of
Support Vector Machines, Ensemble Empirical Mode Decomposition and Partial
Autocorrelation Function.

Despite being relatively new, the field of Deep Learning has attracted a lot of
interest in the past few years. A very thorough review of the entire history of
developments that led the field to its current state can be found in
\cite{schmidhuber2015deep}, while a higher focus on the novelties from the last
decade is given in \cite{arel2010deep}. We proceed to a review of the
applications of Deep Learning to Time-Series data.

\subsubsection{Classification}

The task of Classification of any type of data has benefited by the advent of
CNNs. Previously existing methods for classification
generally relied on the usage of domain specific features normally crafted
manually by human experts. Finding the best features was the subject of a lot
of research and the performance of the classifier was heavily dependent on
their quality. The advantage of CNNs is that they can learn such features by
themselves, reducing the need for human experts \cite{langkvist2014modeling}.

An example of the application of such unsupervised feature learning for the
classification of audio signals is presented in \cite{lee2009unsupervised}. In
\cite{abdel2012applying}, the features learned by the CNN are used as input to
a Hidden Markov Model, achieving a drop at the error rate of over 10\%. The
application of CNNs in these works presuppose the constraint that the
Time-Series is composed of only one channel. An architecture that solves this
constraint is presented in \cite{zheng2014time}.

In \cite{gao2015deep} the performance of CNNs is compared with that of LSTM for
the classification of Visual and Haptic Data in a robotics setting, and in
\cite{jiang2015human} the signals produced by wearable sensors are transformed
into images so that Deep CNNs can be used for classification.

Relevant to Tiled CNNs was the development of Independent Component Analysis
(ICA) \cite{hyvarinen2000independent}. Several alternative methods for
calculating independent components can be found in the literature (e.g.,
\cite{gao2006non}, \cite{suzuki2002fast} or \cite{hyvarinen1999fast}).
Tiled CNNs are normally trained with a
variation of such technique that looses the assumption that each component is
statistically independent and tries to find a {\em topographic order} between
them: the Topographic ICA \cite{hyvarinen2001topographic}.

\subsubsection{Forecasting}

Several different Deep Learning approaches can be found in the literature for
performing Forecasting tasks. For example, Deep Belief Networks are used in the
work of \cite{kuremoto2014time} along with RBM.
\cite{turner2014time} also compares the performance of Deep Belief Networks
with that of Stacked Denoising Autoencoders. This last type of network is also
employed by \cite{romeu2013time} to predict the temperature of an indoor
environment. Another application of Time-Series forecasting can be found in
\cite{lv2015traffic}, which uses Stacked Autoencoders to predict the flow of
traffic from a Big Data dataset.

A popular application to the task of Time-Series prediction is on Weather
Forecasting. In \cite{liu2014deep}, some preliminary predictions on weather data
provided by The Hong Kong Observatory are made through the usage of Stacked
Autoencoders. In a follow up work, the authors use similar ideas to perform
predictions on Big Data \cite{liu2015deep}. Instead of Autoencoders,
\cite{grover2015deep} uses Deep Belief Networks for constructing a hybrid model
in which the ANN models the joint distribution between the weather predictors
variables.

\subsubsection{Anomaly Detection} \label{secRelatedWorkdAnomalyDetect}

Work applying Deep Learning techniques to Anomaly Detection detection of
Time-Series data is not very abundant in the literature. It is still difficult
to find works such as \cite{feng2015novel}, that uses Stacked Denoising
Autoencoders to perform Anomaly Detection of trajectories obtained from low
level tracking algorithms.

However, there are many similarities between Anomaly Detection and the previous
two tasks. For example, identifying an anomaly could be transformed into a
Classification task, as was done in \cite{langkvist2012sleep}. Alternatively,
detecting an anomaly could be considered the same as finding regions in the
Time-Series for which the forecasted values are too different from the actual
ones.

\section{Deep Learning for Time-Series Modeling} \label{secDLModeling}

In this section the work presented in \cite{mittelman2015time} is reviewed. As
discussed above, FCNs are a modification of the CNN architecture that, as
required by some Time-Series related problems, allows for the input and output
signals to have the same dimensions.

Mittelman \cite{mittelman2015time} argues that the architecture of the FCN
resembles the application of
a wavelet transform, and that for this reason, it can present strong variations
when the input signal is subject to small translations. To solve this problem,
and inspired by the undecimeated wavelet transform, which is translation
invariant, they propose the Undecimated Fully Convolutional Neural Network
(UFCNN), also translation invariant.

The only difference between an FCN and an UFCNN is that the
UFCNN removes both the upsampling and pooling operators from the FCN
architecture. Instead, the ``filters at the $l^{th}$ resolution level are
upsampled by a factor of $2^{l-1}$ along the time dimension''. See Figure
\ref{figUFCNN} for a graphical representation of the proposed architecture.

The performance of the UFCNN is tested in three different experiments.
In the first experiment, ``2000 training sequences, 50 validation sequences
and 50 testing sequences, each of length 5000 time-steps'' are automatically
generated by a probabilistic algorithm. The values of the Time-Series represent
the position of a target object moving inside a bounded square. The performance
of the UFCNN in estimating the position of the object at each time-step is
compared to that of a FCN, a LSTM, and a RNN, and the UFCNN does perform better
in most cases.

In a second experiment, the ``MUSE'' and ``NOTTINGHAM''
datasets\footnote{available at http://www-etud.iro.umontreal.ca/~boulanni/icml2012}
area used. The goal is to predict the values of the Time-Series in the next
time-step. In both cases, the UFCNN outperforms the competing networks: a RNN, a
Hessian-Free optimization-RNN \cite{martens2011learning}, and a LSTM.

Finally, the third experiment uses a trading
dataset\footnote{available at http://www.circulumvite.com/home/trading-competition},
where the goal is, given only information about the past, to predict the set of
actions that would ``maximize the profit''. In a comparison to a RNN, the UFCNN
again yielded the best results.

\section{Deep Learning for Time-Series Classification} \label{secDLClassification}

Wang and Oates \cite{wang2015encoding} presented an approach for Time-Series
Classification using CNN-like networks.
In order to benefit from the high accuracy that CNNs
have achieved in the past few years on image classification tasks, the authors
propose the idea of transforming a Time-Series into an image.

Two approaches are presented. The first one generates a Gramian Angular Field
(GAF), while the second generates a Markov Transition Field (MTF). In both
cases, the generation of the images increases the size of the Time-Series,
making the images potentially prohibitively large. The authors therefore propose
strategies to reduce their size without loosing too much information. Finally,
the two types of images are combined in a two-channel image that is then used
to produce better results than those achieved when using each image separately.
In the next sections, GAF and MTF are described.

In the equations below, we suppose that $m = 1$. The Time-Series is therefore
composed by only real-valued observations, such that referring to
$x^{(i)} \in X$ is the same as referring to $\textbf{x}^{(i)} \in X$.

\subsection{Gramian Angular Field}

The first step on generating a GAF is to rescale the entire Time-Series into
values between $[-1, 1]$. In the equation \ref{eqGAFRescale}, $max(X)$ and
$min(X)$ represent the maximum and minimum real-values present in the
Time-Series $X$:

\begin{equation}
\label{eqGAFRescale}
\tilde{x}^{(i)} = \frac{(x^{(i)} - max(X)) + (x^{(i)} - max(X))}{max(X) - min(X)}
\end{equation}

The next step is to recode the newly created Time-Series $\tilde{X}$ into polar
coordinates. The angle is encoded by $x^{(i)}$ and the radius is encoded by the
the time stamp $i$.

Notice that, because the values $x^{(i)}$ were rescaled, no information is lost
by the usage of $arccos(\tilde{x}^{(i)})$ in \ref{eqGAFPolarize}.

\begin{equation}
\label{eqGAFPolarize}
\begin{cases}
\phi = arccos(\tilde{x}^{(i)}), &
    -1 \leq \tilde{x}^{(i)} \leq 1, \tilde{x}^{(i)} \in \tilde{X} \\
r = \frac{i}{N}, & i \in \mathbb{N}
\end{cases}
\end{equation}

Finally, the GAF is defined as follows:

\begin{equation}
\label{eqGAFMatrix}
G =
 \begin{bmatrix}
  cos(\phi_1 + \phi_1) & \cdots & cos(\phi_1 + \phi_n) \\
  cos(\phi_2 + \phi_1) & \cdots & cos(\phi_2 + \phi_n) \\
  \vdots  & \ddots  & \vdots  \\
  cos(\phi_n + \phi_1) & \cdots & cos(\phi_n + \phi_n)
 \end{bmatrix}
\end{equation}

Here, some information is lost by the fact that $\phi$ no more belongs to the
interval $[0, \pi]$. When trying to recover the Time-Series from the image,
there may be some errors introduced.

\subsection{Markov Transition Field}

The creation of the Markov Transition Field is based on the ideas proposed in
\cite{campanharo2011duality} for the definition of the so-called Markov
Transition Matrix (MTM).

For a Time-Series $X$, the first step is defining $Q$ quantile bins. Each
$x^{(i)}$ is then assigned to the corresponding bin $q_j$. The Markov
Transition Matrix is the matrix $W$ composed by elements $w_{ij}$ such that
$\sum_j{w_{ij}} = 1$ and $w_{ij}$ corresponds to the normalized ``frequency with
which a point in the quantile $q_j$ is followed by a point in the quantile
$q_i$.'' This is a $Q \times Q$ matrix.

The MTF is the $n \times n$ matrix $M$. Each pixel of $M$ contains a value from
$W$. The value in the pixel $ij$ is the likelihood (as calculated when
constructing $W$) of going from the bin in which the pixel $i$ is to the bin in
which the pixel $j$ is:


\begin{equation}
\label{eqMTFMatrix}
M =
 \begin{bmatrix}
 w_{ij | x_1 \in q_i, x_1 \in q_j} & \cdots & w_{ij | x_1 \in q_i, x_n \in q_j} \\
 w_{ij | x_2 \in q_i, x_1 \in q_j} & \cdots & w_{ij | x_2 \in q_i, x_n \in q_j} \\
  \vdots  & \ddots  & \vdots  \\
 w_{ij | x_n \in q_i, x_1 \in q_j} & \cdots & w_{ij | x_n \in q_i, x_n \in q_j}
 \end{bmatrix}
\end{equation}

\subsection{Performing Classification with the Generated Images}

The authors use Tiled CNNs to perform classifications using the images. In the
reported experiments, both methods are assessed separately in 12 ``hard''
datasets ``on which the classification error rate is above 0.1 with the
state-of-the-art SAX-BoP approach'' \cite{lin2012rotation}, which are
{\em 50Words}, {\em Adiac}, {\em Beef}, {\em Coffee}, {\em ECG200},
{\em Face (all)}, {\em Lightning-2}, {\em Lightning-7}, {\em OliveOil},
{\em OSU Leaf}, {\em Swedish Leaf} and {\em Yoga} \cite{UCRArchive}.
The authors then
suggest the usage of both methods as ``colors'' of the images. The performance
of the resulting classifier is competitive against many of the state-of-the-art
classifiers, which are also reported by the authors.

\section{Deep Learning for Time-Series Anomaly Detection} \label{secDLAnomalyDetect}

Anomaly Detection
can be easily transformed into a task where the goal is to model the
Time-Series and, given this model, find regions where the predicted values are
too different from the actual ones (or, in other words, where the probability
of the observed region is too low). This is the idea implemented by the paper
reviewed in this section \cite{malhotra23long}.

Through the learned model, not all $m$ input variables need to be predicted. The
learned model predicts, at any given time-step, $l$ vectors with $d$ input
variables, where $1 \leq d \leq m$.

The modeling of the Time-Series is done through the application of a Stacked
LSTM architecture. The network has $m$ input neurons (one for each input
variable) and $d \times l$ output neurons (one neuron for each one of the $d$
predicted variables of the $l$ vectors that are predicted at a time-step). The
hidden layers are composed by LSTM units that are ``fully connected through
recurrent connections''. Additionally, any hidden unit is fully connected to
all units in the hidden layer above it.
Figure \ref{figLSTMAD} sketches the proposed architecture.

For each one of the predictions and each one of the $d$ predicted variables,
a prediction error is calculated. The errors are then used to fit a
multivariate Guassian distribution, and a probability $p^{(t)}$ is assigned to
each observation. Any observation whose probability $p^{(t)} < \tau$ is
treated as an anomaly.

The approach was tested in four real-world datasets. One of them, called
{\em Multi-sensor engine data} is not publicly available. The other three
datasets ({\em Electrocardiograms (ECGs)}, {\em Space Suttle Marotta valve
time series}, and {\em Power demand dataset}) are available for
download\footnote{available at http://www.cs.ucr.edu/~eamonn/discords/}.
The results demonstrated a
significant improvement in capturing long-term dependencies when compared to
simpler RNN-based implementations.

\section{Conclusion} \label{secConclusion}

When applying Deep Learning, one seeks to stack several independent neural
network layers that, working together, produce better results than the already
existing shallow structures. In this paper, we have reviewed some of these
modules, as well the recent work that has been done by using them, found in
the literature. Additionally, we have discussed some of the main tasks normally
performed when manipulating Time-Series data using deep neural network
structures.

Finally, a more specific focus was given on one work performing each one of
these tasks. Employing Deep Learning to Time-Series analysis has yielded
results in these cases that are better than the previously existing techniques,
which is an evidence that this is a promising field for improvement.

\subsubsection*{Acknowledgments.}
I would like to thank Ahmed Sheraz and Mohsin Munir for their guidance and
contribution to this paper.

\bibliographystyle{splncs}

\bibliography{seminar}

\end{document}